# Data Classification Using the Dempster-Shafer Method

Qi Chen[a], Amanda Whitbrook[a], Uwe Aickelin[a] and Chris Roadknight[ab]

[a]Intelligent Modelling and Analysis Research Group, School of Computer Science, University of Nottingham, Jubilee Campus, Wollaton Road, Nottingham, NG8 1BB, U.K.
Email: <qxc, amw, uxa, cmr>@cs.nott.ac.uk, Telephone: +441158466554, Fax: +441158467877.

[b] Corresponding author

**Abstract**

In this paper, the Dempster-Shafer method is employed as the theoretical basis for creating data classification systems. Testing is carried out using three popular (multiple attribute) benchmark datasets that have two, three and four classes. In each case, a subset of the available data is used for training to establish thresholds, limits or likelihoods of class membership for each attribute, and hence create mass functions that establish probability of class membership for each attribute of the test data. Classification of each data item is achieved by combination of these probabilities via Dempster's Rule of Combination.  Results for the first two datasets show extremely high classification accuracy that is competitive with other popular methods. The third dataset is non-numerical and difficult to classify, but good results can be achieved provided the system and mass functions are designed carefully and the right attributes are chosen for combination. In all cases the Dempster-Shafer method provides comparable performance to other more popular algorithms, but the overhead of generating accurate mass functions increases the complexity with the addition of new attributes. Overall, the results suggest that the D-S approach provides a suitable framework for the design of classification systems and that automating the mass function design and calculation would increase the viability of the algorithm for complex classification problems.

**Keywords**

Dempster-Shafer theory, data classification, Dempster's rule of combination.

## 1. Introduction

The ability to group complex data into a finite number of classes is important in data mining, and means that more useful decisions can be made based on the available information. For example, within the field of medical diagnosis, it is essential to utilise methods that can accurately differentiate between anomalous and normal data.

Systems that categorize data in this way are usually referred to as *classifiers*, and fall into two distinct types; statistical and Artificial Intelligence (AI) methods. Examples of AI methods include fuzzy classifiers [13], support vector machines [14], and *k*-nearest neighbour techniques [15]. Here, Dempster's rule of combination (DRC), a statistical method, is considered as a tool for classification. DRC is a generalization of a special case of Bayes' theorem, concerned with combining *independent* sets of probability assignments to form a single one. The method has previously been used for combining results from several different classifiers [16, 17]. However, in this research the intention is to develop classification systems based on the D-S theory only.

The chief aims here are to describe the use of the Dempster-Shafer (D-S) theory as a framework for creating classifier systems, test the systems on three benchmark datasets, and compare the results with those for other techniques. Organization of the paper is as follows; Section 2 discusses the fundamentals of the D-S theory, its advantages and disadvantages, and its previous use as a classification tool in the literature. Section 3 describes the basic general principles for classifier system design using the theory. Sections 4, 5, and 6 provide information about the three standard benchmark problems, the Wisconsin Breast Cancer Dataset (WBCD) [1], the Iris Plant Dataset (IPD) [1], and the Duke Outage Dataset (DOD) [18-26] respectively. These sections also detail the individual D-S system architectures used for classifying each dataset, and present their results. Section 7 concludes the paper.

## 2. The Dempster-Shafer (D-S) theory

D-S is a mathematical theory of evidence based on belief functions and plausible reasoning. It was introduced in the 1960's as a mechanism for reasoning under epistemic (knowledge) uncertainty by Arthur Dempster [2], and developed in the 1970's by Glenn Shafer [3]. The theory contains several distinct models, for example the Transferable Belief Model (TBM), which obtains degrees of belief for one question from subjective probabilities for a related question. Also available is Dempster's Rule of Combination (DRC), which seeks to combine probabilities when they are based on independent items of evidence.

The D-S theory uses the same fundamental domain as probability theory but is relevant to those situations where non-random uncertainty is present, for example where data are collected from partially reliable sources. It is hence suited to problems where the actual state of affairs is known only to belong to some subset of possible states.

The TBM model, which incorporates some essential mathematical terminologies underpinning the D-S theory, is presented in Section 2.1, and Section 2.2 introduces DRC, the part of the theory of direct relevance to the classification systems described here. The advantages and disadvantages (in particular the computational complexity problem and conflict management problem) of D-S are discussed in Section 2.3. Section 2.4 provides an overview of past work in this field.

## 2.1. Basic mathematical terminology and the TBM

The D-S theory begins by assuming a *frame of discernment* ($\Theta$), which is a finite set of mutually exclusive propositions and hypotheses (alternatives) about some problem domain. It is the set of all states under consideration. For example, when diagnosing a patient, $\Theta$ would be the set consisting of all possible diseases. The *power set* $2^\Theta$ is the set of all possible sub-sets of $\Theta$ including the empty set $\Phi$. For example, if:

$$\Theta = \{a, b\}, \tag{1}$$

then

$$2^\Theta = \{\phi, \{a\}, \{b\}, \Theta\}. \tag{2}$$

The individual elements of the power set represent propositions in the domain that may be of interest. For example, the proposition "the disease is infectious" gives rise to the set of elements of $\Theta$ that are infectious and contains all and only the states in which that proposition is true.

The theory of evidence assigns a *mass value m* between 0 and 1 to each subset of the power set. This can be expressed mathematically as:

$$m : 2^\Theta \to [0, 1] \tag{3}$$

The function (3) is called the *mass function* (or sometimes the *basic probability assignment*) whenever it verifies two axioms: First, the mass of the empty set must be zero:

$$m(\phi) = 0, \tag{4}$$

and second, the masses of the remaining members of the power set must sum to 1:

$$\sum_{A \subseteq \Theta} m(A) = 1 \cdot \tag{5}$$

The quantity $m(A)$ is the measure of the probability that is committed exactly to $A$ [3]. In other words, $m(A)$ expresses the proportion of available evidence that supports the claim that the actual state belongs to $A$ but not to any subset of $A$.

Given mass assignments for the power set, the upper and lower bounds of a probability interval can be determined since these are bounded by two measures that can be calculated from the mass, the *degree of belief (bel)* and the *degree of plausibility (pl)*. The degree of belief function of a proposition $A$, $bel(A)$, sums the mass values of all the non-empty subsets of $A$:

$$bel(A) = \sum_{B \subseteq A} m(B). \tag{6}$$

The degree of plausibility function of $A$, $pl(A)$, sums the masses of all the sets that intersect $A$, i.e. it takes into account all the elements related to $A$ (either supported by evidence or unknown):

$$pl(A) = \sum_{B \cap A \neq \phi} m(B). \tag{7}$$

The degree of belief function and degree of plausibility function are related to each other as follows:

$$Pl(A) = 1 - bel(\neg A). \tag{8}$$

For the subset *A*, *pl(A)* and *bel(A)* represent upper and lower bounds of the probability interval respectively, and the interval [*bel(A)*, *pl(A)*] represents the probability range or uncertainty. The relationships between *bel* value, *pl* value and uncertainty are shown graphically in Figure 1.

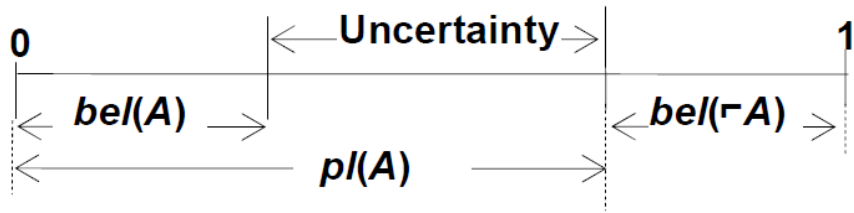

Fig. 1. The uncertainty interval for a given hypothesis [4].

Note that it is always true that:

$$m(A) \leq bel(A) \leq pl(A). \tag{9}$$

As an example, consider an object that may be coloured in only one of three colours, red, green or yellow. If the mass values for the hypotheses in the power set are as given in column 2, then the degree of belief and degree of plausibility functions may be derived.

Table 1 Example mass, belief and plausibility values

| Hypotheses | Mass | Belief | Plausibility |
|---|---|---|---|
| 1  Null | 0 | 0 | 0 |
| 2  Red | 0.24 | 0.24 | 0.45 |
| 3  Green | 0.16 | 0.16 | 0.37 |
| 4  Yellow | 0.33 | 0.33 | 0.57 |
| 5  Red or green | 0.03 | 0.43 | 0.67 |
| 6  Red or yellow | 0.06 | 0.63 | 0.84 |
| 7  Green or yellow | 0.06 | 0.55 | 0.76 |
| 8  Any Θ | 0.12 | 1.00 | 1.00 |

The degree of belief values for hypotheses 2, 3 and 4 are the same as the mass values as they contain no subsets. Note that belief and mass values are always equivalent in the case of singleton sets, i.e. sets with only one member. However, the degree of belief value for hypothesis 5 is the sum of the mass values for itself and its subsets (hypotheses 5, and 2 and 3 respectively). The degree of plausibility of hypothesis 2 is 1 minus the degree of belief of hypothesis 7, and the degree of plausibility of hypothesis 5 is 1 minus the degree of belief of hypothesis 4.

## 2.2. Dempster's rule of combination

DRC is concerned with uniting two independent sets of mass functions on a frame of discernment $\Theta$, for example $m_1$ and $m_2$. In this case, the combined mass function of $m_1$ and $m_2$ (the joint mass) would be expressed as $m_{1,2}$ where:

$$m_{1,2}(A) = (m_1 \oplus m_2)(A), \tag{10}$$

and

$$(m_1 \oplus m_2)(A) = \frac{1}{1-K} \sum_{B \cap C = A \neq \phi} m_1(B) m_2(C), \tag{11}$$

where $K$, a measure of the conflict between the two mass sets, is given by:

$$K = \sum_{B \cap C = \phi} m_1(B) m_2(C), \quad K \neq 1. \tag{12}$$

Equation (11) essentially emphasizes the agreement between multiple sources of information and ignores conflicting evidence by using a normalization factor, which is equal to 1-$K$. The normalization factor attributes any mass associated with conflict to the null set, although this can often yield counterintuitive results if the level of conflict is high, see Section 2.3.

As an example DRC calculation, suppose the mass values in the set $M_1$ are $m_1(n) = 0.4$, and $m_1(a) = 0.6$, and the mass values in the set $M_2$ are $m_2(n) = 0.2$, and $m_2(a) = 0.8$, where $n$ represents the hypothesis normal and $a$ represents the hypothesis abnormal. First, $K$ is calculated by multiplying those masses for which the intersect is the empty set, and summing the results:

$$\begin{aligned} K &= (m_1(n) \times m_2(a)) + (m_1(a) \times m_2(n)) \\ &= (0.4 \times 0.8) + (0.6 \times 0.2) = \frac{44}{100}. \end{aligned} \tag{13}$$

The joint masses are then derived by multiplying those masses for which the intersect is the proposition of interest, summing them and multiplying the result by the reciprocal of the normalization factor:

$$(m_1 \oplus m_2)(n) = \frac{100}{56} \times m_1(n) m_2(n) = \frac{1}{7}, \tag{14}$$

and

$$(m_1 \oplus m_2)(a) = \frac{100}{56} \times m_1(a) m_2(a) = \frac{6}{7}. \tag{15}$$

## 2.3. Advantages and disadvantages of D-S

The systems described in this paper are all based on the theory presented in Sections 2.1 and 2.2, but D-S-based systems have a great deal of scope and flexibility as regards to system design, which means that classifiers can be created that are highly suited for solving any given problem. In particular, there are no fixed rules regarding how the mass functions should be constructed or how the data combination should be organized. One can build mass functions that are as simple or as complex as desired, and DRC can be applied many times using different strategies. This is a distinct advantage in that it allows the creation of systems tailored towards a given problem domain, but it may be argued that it is also a disadvantage, since it does not permit the generalization of results to all domains.

However, a definite advantage of D-S is that no *a priori* knowledge is required, making it potentially suitable for classifying previously unseen information. Furthermore, a value for ignorance can be expressed, providing information on the uncertainty of a situation. In contrast, Bayesian inference requires some *a priori* knowledge and is unable to assign a probability to ignorance. A Bayesian approach is therefore not always suitable for classification since pre-existing knowledge may not always be provided, for example, if the aim is to detect previously unseen network attacks in computer systems.

There are two major problems associated with D-S, namely:

- The computational complexity problem
- The conflicting beliefs management problem

The computational complexity increases exponentially with the number of elements in the frame of discernment ($\Theta$). If there are $n$ elements in $\Theta$, there will be up to $2^{n-1}$ focal elements for the mass function, and the combination of two mass functions would require the computation of up to $2^n$ intersections. To overcome this, various algorithms have been suggested, such as [5] and [6], which reduce the number of focal elements in the mass functions involved. However, for the datasets considered here, computational complexity is low, as the frame of discernment consists of only two elements for the WBCD, three elements for the ID and four for the DOD dataset.

Zadeh (1986) [12] was the first to describe the conflicting beliefs management problem. As its name suggests, this occurs when collected mass values from different information sources conflict with each other. When this is the case, attempts to combine evidence using DRC are likely to produce counterintuitive results. For example, consider the case where a car window has been broken and there are three suspects Jon, Mary, and Mike, and two witnesses, W1 and W2. W1 assigns a mass value of 0.9 to "Jon is guilty" and a mass value of 0.1 to "Mary is guilty". However, W2 assigns a mass value of 0.9 to "Mike is guilty" and a mass value of 0.1 to "Mary is guilty". Applying the DRC returns a value of 0.99 for $K$, which yields a value of 1 for "Mary is guilty". This is clearly counterintuitive since both witnesses assigned very small mass values to this hypothesis. The conflicting beliefs management problem is only a cause for concern when there are more than two classes, so the WBCD dataset used here presents no potential problem. Furthermore, the mass functions used with other two datasets are selected so that any conflicting beliefs are reduced (see Sections 5.2 and 6.2). This is possible since the problem is caused by conflicting mass values, not mass functions, so one can design mass functions and DRC combination strategies that minimize the problem. Some alternative combination rules that attempt to reduce the conflicting beliefs management problem have also been proposed, as in [7] and [8], but none have yet been accepted as a standard method.

## 2.4. Review of D-S applications

The D-S theory has previously been shown to be a powerful combination tool, but to date most of the research effort has been directed towards using it to unite the results from a number of separate classification techniques. For example, in [30] the results from a Bayesian network classifier and a fuzzy logic-based classifier are combined and in [31] the D-S theory is used in conjunction with a neural network methodology and applied to a fault diagnosis problem in induction motors. The DRC acts as a data fusion tool, i.e. eight faulty conditions are first classified using the neutral network and the classification information is then converted to mass function assignments. These are then combined using DRC, which reduces the diagnostic uncertainty. Al-Ani and Deriche [32] also propose a classifier combination method based on the D-S approach. They propose that the success of the D-S methodology lies in its powerful ability to combine evidence measures from multiple classifiers. In other words, when the results of several classifiers are combined, the effects of their individual limitations as classifiers are significantly reduced. Valente and Hermansky [33] also suggest a DRC methodology that combines the outputs from various neural network classifiers, but in their work it is applied to a multi-stream speech recognition problem.

As mentioned previously, the work here differs from the above approaches in that it is concerned with classification using the D-S theory alone; no other categorization techniques are employed at any stage in the classification process. This perspective is fairly novel as other works concerned with the D-S theory as a single classifier have mostly focused on adapting its methodology. For example, Parikh *et al.* [34] present a new method of implementing D-S for condition monitoring and fault diagnosis, using a predictive accuracy rate for the mass functions. The authors claim that this architecture performs better than traditional mass assignment techniques as it avoids the conflicting beliefs assignment problem.

In other D-S related work, Chen and Venkataramanan [35] show that Bayesian inference requires much more information than the D-S theory, for example *a priori* and conditional probabilities. They postulate that the D-S method is tolerant of trusted but inaccurate evidence as long as most of the evidence is accurate.

## 3. The application of D-S to data classification

As discussed in Section 2.3, the D-S theory provides a general framework for creating classifier systems. This framework can be expressed as a series of steps that must be undertaken namely:

1. Define the frame of discernment ($\Theta$). This is the set of all possible hypotheses related to the given dataset and identifies the classes to which the data must be assigned.

2. Determine which data attributes are important for establishing class membership and discard the others. In general, the frame of discernment and the selected attributes (their number and their data types) will provide loose guidelines for designing mass functions and the structure of the DRC combinations.
3. Examine the selected attributes and their data values within a subset of the data in order to design mass functions for each attribute. These functions will be used to assign mass values to the corresponding hypotheses based on the attribute values of the test data.
4. Design a DRC combination strategy based on the data structure. A single application of DRC combines the mass values of each attribute for each data item, but many applications can be used, and DRC can also be used to combine the results of previous applications.
5. Following combination, select a rule that converts the result to a decision. Several may be used on different steps, but the final one ultimately classifies the data.

These steps are best illustrated by example. Here, the D-S methodology is implemented on three standard benchmark problems WBCD, ID, and DOD (see Sections 4, 5, and 6 respectively). In each case, the system uses training data (a subset of the original dataset) to derive mass functions for each attribute, and hence determine probability of class membership for each attribute when the test data (data to be classified) is used. The mass functions used for the WBCD are sigmoid functions, based on training data threshold values. For the ID dataset, the mass functions are related to training data boundary values for each attribute and each class, and for the DOD data, the mass functions use the likelihood measure.

A DRC scheme may be a simple one-step process as with the WBCD (see Figure 2 and Section 4). Alternatively, it may be more complex, for example a two-step process as in the case of the ID dataset (see Section 5), or a branching multi-step process as with the DOD (see Section 6). Further details on construction of suitable architectures may be found in [36] and [37].

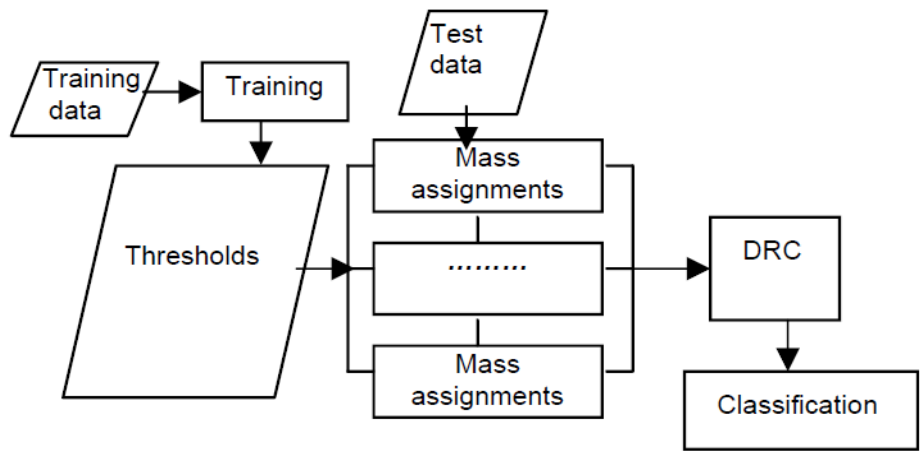

Fig. 2. Data flow diagram for a one-step D-S classification system .

## 4. Experiments with the WBCD

### 4.1. Description of the data

The WBCD is a standard benchmark dataset of the UCI Machine Learning Repository [1]. It contains 699 data items, 241 of which are malignant (abnormal data), and 458 of which are benign (normal data). The data consists of nine attributes (A, B, C, D, E, F, G, H and I) that are all normalised integers in the range 1 to 10, with smaller values tending to indicate normality and larger values tending to indicate abnormality. The biological meanings of these attributes are summarized in Table 2 below.

**Table 2 Attributes in the WBCD dataset**

| Attribute | Biological Interpretation |
|---|---|
| A | Clump thickness |
| B | Uniformity of cell size |
| C | Uniformity of cell shape |
| D | Marginal adhesion |
| E | Single epithelial cell size |
| F | Bare nuclei |
| G | Bland chromatin |
| H | Normal nucleoli |

The WBCD is selected as it is a popular dataset with published performance results for a number of other methods, which facilitates easy comparison of the D-S approach with other algorithms, see Section 4.3. In particular, it is worth noting that there are 16 data items that have one missing (i.e. unavailable) attribute value. The D-S based classification system described here has the ability to cope with this problem by omitting, i.e. not combining, the missing attribute values of the corresponding data items. This is an advantage over other approaches, for example [9] and [10], which would have to exclude the 16 items with missing values. In addition, with the D-S theory it is possible to attempt to classify data items based on single attributes, combinations of attributes or using all attribute values, and hence determine which combinations work best.

### 4.2. The classification approach

For the WBCD, the frame of discernment is {normal, abnormal}, with normal defining a benign item and abnormal defining a malignant item. Ten-fold cross validation is used, i.e. the dataset is randomly divided into ten subsets of approximately equal size (one subset size is either 69 or 70), and for each run the data of one subset is used as test data, and the data of the other nine subsets is used as training data.

Training consists of obtaining the modified median thresholds $t$ to build the mass functions. The full dataset size is 699, so the training data size is either 630 or 629. For each attribute, the training data values are ordered from small to large, and since the proportional distribution of the WBCD is 65.5% : 34.5% (normal : abnormal), the 413$^{th}$ value of each attribute is chosen as the modified median threshold. (Note that if the training data size is 629, the 412$^{th}$ value is used.) A general assumption is that lower-valued items tend to represent normal data, so the mass functions for each attribute can be modelled using sigmoid functions:

$$m(normal) = (1 + e^{(v-t)})^{-1},$$
$$m(abnormal) = 1 - m(normal),$$
(16)

where $v$ is the value of the test data item for that attribute. Figure 3 shows the shape of the mass function $m(normal)$ when a sample threshold value of 5 is used. Note that for all of the datasets considered here, all data items are integers and hence the functions consist of discrete values only.

The mass functions are used to assign mass values to each attribute for each item in the test data, and these are then combined using DRC to obtain overall normal and abnormal mass values. For a given data item, if the mass value of the abnormal hypothesis is larger than that of the normal hypothesis, then that item is classified as abnormal, otherwise it is classified as normal. The architecture is represented schematically in Figure 4.

Classification accuracy $C_a$ is used to judge the quality of the results. This is given by:

$$C_a = \frac{N_c}{N_t},$$
(17)

where $N_c$ is the number of correctly classified items and $N_t$ is the total number of data items. Classification is carried out using DRC on all the attributes, on attributes A, D, and H only, on attributes B, C, and F only and using each of the attributes separately, i.e. twelve different classification scenarios are carried out. This is to test which attribute combinations give rise to the best performance, and whether increasing the number of attributes improves the classification accuracy.

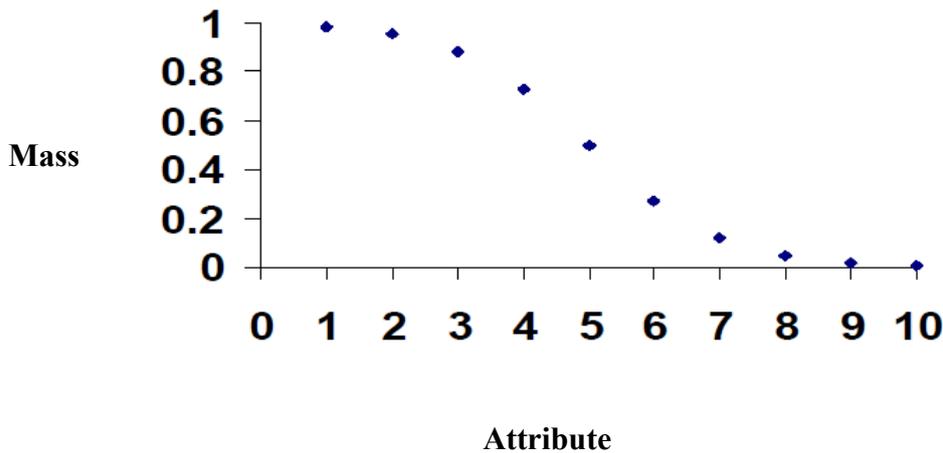

Fig. 3. The shape of the mass function *m*(*normal*) for the WBCD dataset.

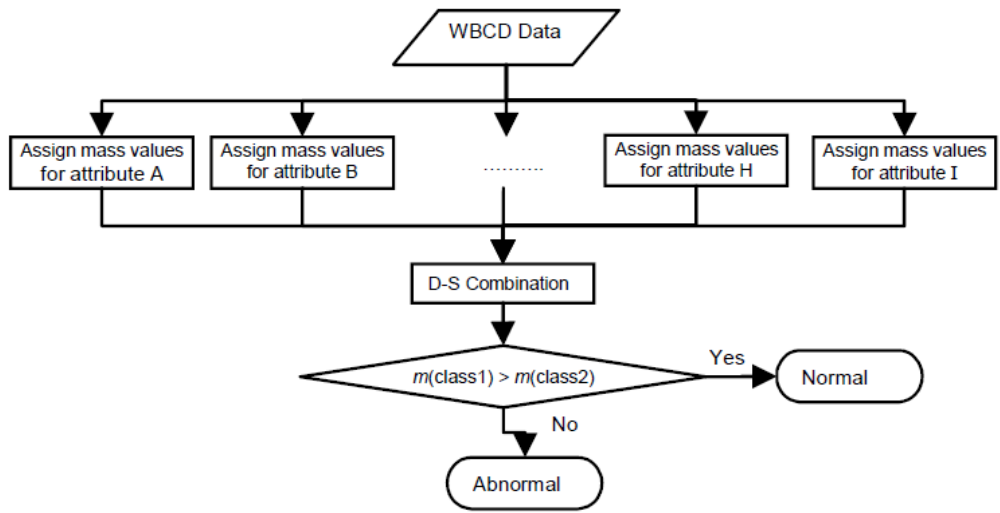

Fig. 4. System flowchart for classification of the WBCD dataset.

### 4.3. Experimental results for the WBCD

Classification accuracy values are plotted for each of the twelve scenarios in Figure 5 below. The plot shows that single attributes A ($C_a$ = 86.0%), D ($C_a$ = 85.7%) and I ($C_a$ = 79.3%) give rise to the poorest performances when using only one attribute at a time. However, when combining these three attributes using DRC, a $C_a$ value of 90.0% is achieved. In addition, DRC combination of attributes B (individual $C_a$ = 92.7%), C (individual $C_a$ = 92.1%), and F (individual $C_a$ = 91.3%), yields a $C_a$ value of 95.7%, i.e. a value greater than that exhibited by any of the constituent attributes. Moreover, the 10 fold cross validated result of using all nine attributes ($C_a$ = 97.6%) is better than any other combination.

    The technique used here and the classification rate of **97.6%** proves highly competitive with other algorithms, for example [9] uses a generalized rank nearest neighbour rule and achieves a classification rate of **96.2%**. Nauk and Kruse [10] use a fuzzy classification method, with a best classification rate of **96.7%**. However, both methods do not take account of the 16 WBCD data items that have missing attribute values. In addition, application of the Naive Bayes classifier approach described in [38], which permits inclusion of the 16 missing data items achieves a **95.0%** success rate. Naive Bayes classification requires no *a priori* knowledge and, whilst it makes the assumption that features are independent, it is often competitivel with more sophisticated classifiers.

The WBCD results show that combining attributes improves classification accuracy consistently when this particular D-S system is used. Indeed, it seems that a few badly-chosen attributes do not influence the results negatively, as long as most are suitable. The success of the D-S approach with this dataset suggests that it could be highly amenable for solving other real-world problems.

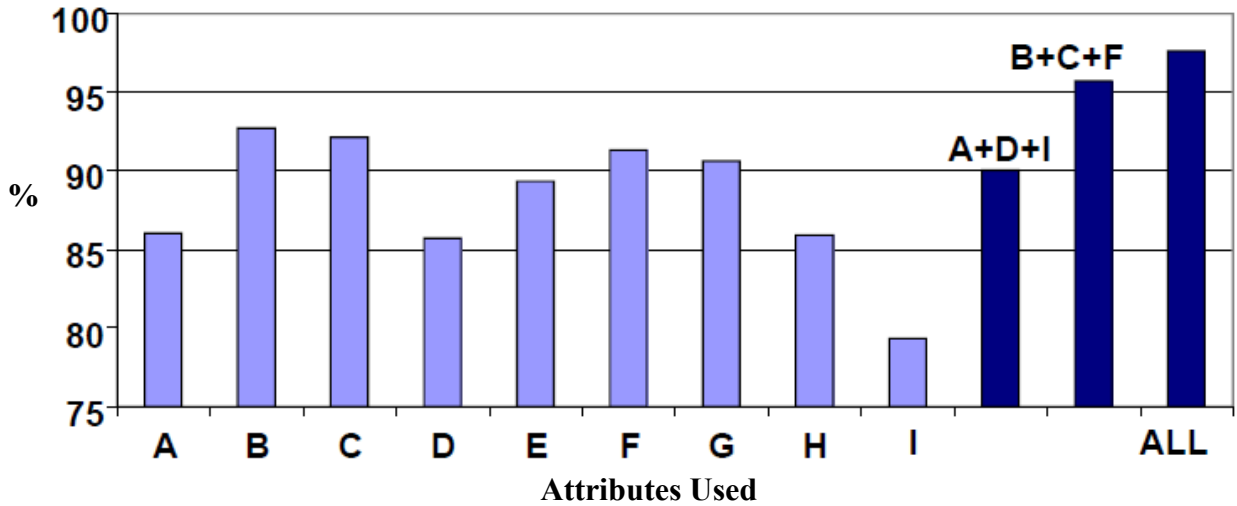

Fig. 5.   Classification accuracies for the WBCD using different attribute combinations.

## 5. Experiments with the Iris Plant dataset

### 5.1. Description of the data

The IPD is another standard benchmark problem of UCI datasets [1]. It has fewer attributes and more classes than the WBCD and thus presents an ideal choice for testing the robustness of the D-S approach when applied to such problems. The dataset has 150 data items with the following four numeric attributes; sepal length, sepal width, petal length, and petal width (all in cm).  The classes are the plant types, namely; *Iris Setosa, Iris Versicolour,* or *Iris Virginica*, with each class containing 50 instances.

### 5.2. The classification approach

The frame of discernment for the IPD is {Setosa, Versicolour, Virginica}, and the chosen initial classification step allows for seven possible hypotheses:

- {Setosa} = class 1
- {Versicolour} = class 2
- {Virginica} = class 3
- {Setosa, Versicolour}
- {Setosa, Virginica}
- {Versicolour, Virginica}
- {Setosa, Versicolour, Virginica}

As for the WBCD, ten-fold cross validation is used, i.e. the dataset is randomly divided into ten subsets of equal size, with nine out of ten subsets comprising training data, and the remaining subset being used as test data. Again, the training data is used to build the mass functions, but since there are three classes, this is a slightly more complex process than before.

First, the maximum and minimum values for each class based on one attribute of the training data are found, and the overlap between these is used to obtain boundary information for each class, see Figure 6. This initial step provides a rough method for grouping the data, although, as can be seen from the diagram, some items may not be classified into a single class. For example, if an attribute value is less than min(class 2), and greater than or equal to min(class 1), the data item must belong to class 1. However, if the value is less than min(class 3), and greater than or equal to min(class 2), the data item could belong to either class 1 or class 2. The process is repeated for each attribute and mass values are assigned in the following way. For data items that are assigned to one class $x$ only:

$$m(class\ x) = 0.9,\ m(\Theta) = 0.1, \tag{18}$$

i. e. the uncertainty $m(\Theta)$ is fixed at 0.1. For data items assigned to two possible classes $x$ and $y$:

$$m(class\ x \cup class\ y) = 0.9,\ m(\Theta) = 0.1. \tag{19}$$

For data items assigned as possibly belonging to all three classes:

$$m(\Theta) = 1, \tag{20}$$

i. e. the uncertainty $m(\Theta)$ is 1.

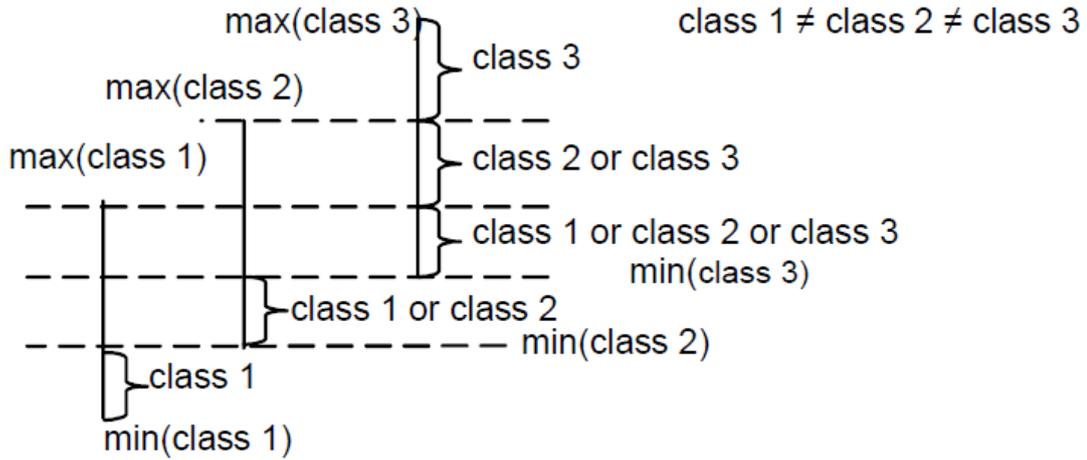

Fig. 6. Calculation of the three-class overlap for one attribute.

The system combines the mass values from all four attributes using DRC, thus producing overall mass values for all hypotheses, and the hypothesis with the highest belief value is used to classify the data item. If the hypothesis does not represent a single class, then a second step is necessary.

In the second step, for each attribute, standard deviations are calculated for each individual class in question and for the union of those classes. The Feature Selection Value $FSV$ is then calculated based on:

$$FSV = \frac{sd(class_1) \times sd(class_2) \times \cdots \times sd(class_n)}{sd(class_1 \cup class_2 \cup \cdots \cup class_n)}, \tag{21}$$

where $n$ is a natural number representing the number of classes. The attribute $a$ that has the smallest $FSV$ is selected, and the absolute difference $d$ (defined in (22)) between the data item's $a$ value and the mean $a$ value is calculated for each class,

$$d_i = |a_i - \bar{a}_i|,\ i = 1, \ldots, n. \tag{22}$$

The data item is classified as belonging to the class $z$ with the smallest $d$ value. The complete classification process is illustrated in a flow chart in Figure 7. Note that the system was coded and executed using the same set-up as the WBCD.

## 5.3. Experimental results for the IPD

Ten runs based on both steps of the approach yields a mean classification accuracy $C_a$ of 95.47%. Table 3 illustrates three out of these ten trials (chosen randomly), and provides detailed error information. The column headings of Table 3 are explained below:

- ID - Data item identification number:
  - 1-50     =     Setosa,
  - 51-100   =     Versicolour,
  - 101-150  =     Virginica.
- Correct (1st) - The number of items correctly classified at the 1$^{st}$ step.
- Errors (1st) - The number of errors caused by the first step.
- Split (1st) - The number of data items not in a single class after the first step.
- Errors (2nd) - The number of errors caused by the second step.

From Table 3 it can be seen that data items 71, 86, 107 and 120 are wrongly classified in all three runs, and in each case the error occurs on the first step. Data item 78 is also wrongly classified in all three runs, but the error occurs on the second step.

In order to analyze the source of error, the results of the 2$^{nd}$ run for data item 86 are examined. Table 4 shows the boundary values used in this run for each class and each attribute. For this data item attribute 1 is 6.0 cm, which suggests membership of either class 2 or class 3, and attribute 2 is 3.4 cm which would indicate that it belongs either to class 1 or class 3. Attribute 3 is 4.5 cm and attribute 4 is 1.6 cm, which both suggest class 2 or class 3 membership. The error occurs following DRC on the first step, i.e. the item is wrongly classified as belonging to class 3. Errors such as this could potentially be reduced by using both steps for all data items, rather than just those that are not placed in single classes. A finer-grained mass model might also lead to better performance; these are subjects for further investigation.

**Table 3 Example experimental results**

1$^{st}$ Run — $C_a$ = 96.6667%

| ID | Correct (1$^{st}$) | Errors (1$^{st}$) | Split (1$^{st}$) | Errors (2$^{nd}$) |
|---|---|---|---|---|
| 1-50 | 50 | 0 | 0 | 0 |
| 51-100 | 35 | 2 (ID = 71, 86) | 13 | ID = 78 |
| 101-150 | 42 | 2 (ID = 107, 120) | 6 | 0 |

2$^{nd}$ Run — $C_a$ = 95.3333%

| ID | Correct (1$^{st}$) | Errors (1$^{st}$) | Split (1$^{st}$) | Errors (2$^{nd}$) |
|---|---|---|---|---|
| 1-50 | 50 | 0 | 0 | 0 |
| 51-100 | 33 | 4 (ID = 51, 71, 84, 86) | 13 | ID = 78 |
| 101-150 | 42 | 2 (ID =107, 120) | 6 | 0 |

3$^{rd}$ Run — $C_a$ = 94.6667%

| ID | Correct (1$^{st}$) | Errors (1$^{st}$) | Split (1$^{st}$) | Errors (2$^{nd}$) |
|---|---|---|---|---|
| 1-50 | 50 | 0 | 0 | 0 |
| 51-100 | 34 | 5 (ID = 51, 57, 71, 84, 86) | 11 | ID = 78 |
| 101-150 | 43 | 2 (ID =107, 120) | 5 | 0 |

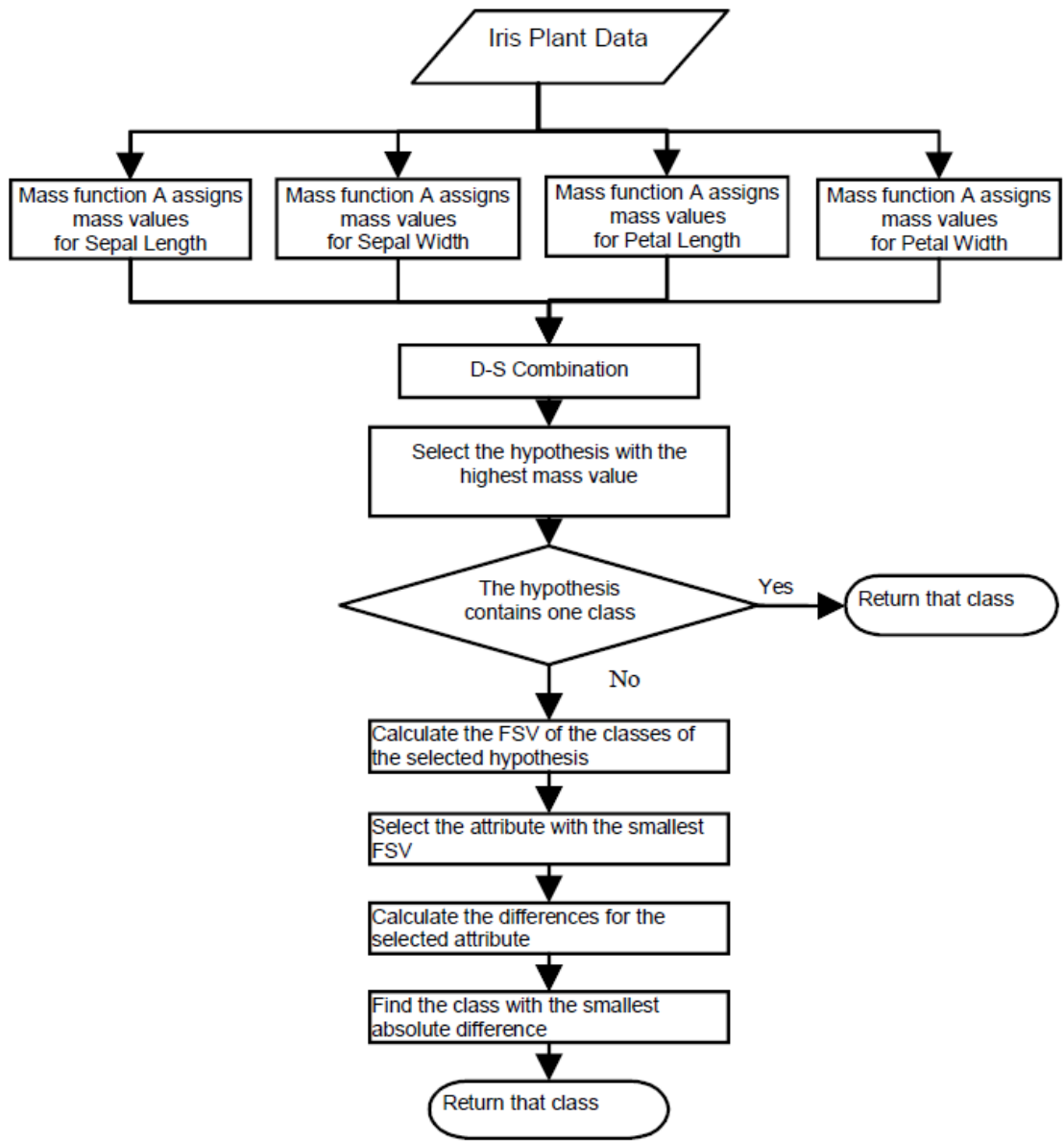

**Fig. 7.** System flowchart for classification of the Iris Plant Dataset.

The results achieved are similar to published values for fuzzy methods, which have $C_a$ values ranging from 94.67% to 97.33% [11], and the results of the Naive Bayes approach, which achieves 96% when ten-fold cross validated. The results suggest that this particular D-S combination system is equivalent to class leading approaches in its ability to classify items with multiple classes and fewer attributes than the WBCD dataset.

**Table 4 Training set boundary values for the 2nd run**

| Attribute | | Class 1 | Class 2 | Class 3 |
|---|---|---|---|---|
| 1 | max | 5.8 | 6.9 | 7.9 |
|   | min | 4.3 | 4.9 | 4.9 |
| 2 | max | 4.4 | 3.3 | 3.8 |
|   | min | 2.3 | 2.0 | 2.2 |
| 3 | max | 1.9 | 5.1 | 6.7 |
|   | min | 1.0 | 3.3 | 4.5 |
| 4 | max | 0.6 | 1.7 | 2.5 |
|   | min | 0.1 | 1   | 1.4 |

# 6. Experiments with the Duke Outage dataset

## 6.1. Description of the data

The Duke Outage Dataset (DOD) is a log of power failures (either complete power loss or a voltage decrease) that occurred in the USA between 1994 and 2002. One of the main reasons for keeping such records is their use as a diagnostic system to help engineers identify the outage cause (for example equipment failure, animal contacts, trees, lightning, etc.) so that the response and restoration procedure can be mounted rapidly and reliably. Here, the dataset is used as a means of testing the application of the D-S methodology on non-numerical data. It is also selected because it has proved difficult to classify accurately in previously published work [19, 20] using different methods, i.e. both good true positive rates and good true negative rates are difficult to achieve for all data classes. This may be partly because the DOD is an imbalanced dataset, with 6.99% of faults caused by lightning, 20.58% due to animals, 29.40% occurring because of fallen trees, and the remaining 43.03% attributed to other causes. There is also a lot of inconsistent data.

There are a total of 33 attributes in each outage record of the DOD, although, based on the statistical significance tests of Xu and Chow [24] and Chow and Taylor [22], only ten of these are considered influential for determining the cause of the fault. These are listed and explained in Table 5.

Table 5 Important attributes in the Duke Outage dataset

| Attribute Name | Example Data | Notes |
| --- | --- | --- |
| Circuit ID Number | 25151202 | ID no. of circuit breaker relay activated |
| Weather | 00 | 00 is code for Fair |
| Year | 1994 | |
| Month | 12 | December |
| Day | 9 | 9th |
| Time of day | 13 | 1pm |
| Phases affected | Z | Only phase Z affected |
| Protective devices activated | 03 | Outage activated a primary fuse, code 03 |
| Cust_off | 1 | Only 1 customer affected |
| Duration | 55 | Outage lasted 55 minutes |

The data is divided into subsets by area, and here only the Clemson Area data is considered.

## 6.2. The classification approach

Any root-cause identification problem can be viewed as a classification problem. For example, here each outage needs to be assigned to a cause class label. To facilitate this, the ten influential attributes listed in Table 5 are reduced to six attributes following [18]. These are:

1. Circuit ID Number (CI)
2. Weather (WE)
3. Season (SE)
4. Time of Day (TD)
5. Phases Affected (PA)
6. Protective Devices Activated (PD)

In other words, Year, Cust_off and Duration are ignored and Month and Day are amalgamated into the new attribute Season. Except for CI, which is numerical, all these attributes are categorical variables; (some example data is provided in Table 6). Since only animal, tree and lightning faults are of interest here, the frame of discernment ($\Theta$) is {Tree, Animal, Lightning, Others}, i.e. the actual outage fault type is used as the class label. This set includes all the possible hypotheses of the system, as a fault is not considered to have multiple causes.

**Table 6 Example Duke Outage data**

| CI | WE | SE | PD | PA | TD | Fault |
|---|---|---|---|---|---|---|
| 25701206 | 10 | Autumn | 4 | 1 | Afternoon | Tree |
| 25061204 | 4 | Spring | 3 | 1 | Afternoon | Tree |
| 25061203 | 0 | Summer | 3 | 1 | Midnight | Tree |
| 25251201 | 1 | Winter | 4 | 1 | Midnight | Animal |
| 25712402 | 4 | Autumn | 3 | 1 | Midnight | Animal |
| 25701206 | 4 | Spring | 3 | 1 | Midnight | Animal |

The D-S system created for this dataset is designed to treat each class separately, i.e. the data is initially analysed three times. First each data item is classed as either tree or non-tree, then as either animal or non-animal, and finally as either lightning or non-lightning. In each case, this is done by combining the mass values from each attribute using DRC. The mass values from the results of the three initial classifications are then combined using DRC. The classification strategy is shown as a flow chart in Figure 8, and has a flow structure based on the Artificial Neural Network (ANN) system described in [20], which proves to be far superior to a system that performs only one set of classification results for membership of the four classes. Furthermore, the architecture of the coding permits the easy addition of other classes if they are needed, i.e. the system can be expanded to $N$ branches for $N$ outage-fault types.

Essentially, the system can be subdivided into the following components:

- Collect the information needed for building the mass functions, i.e. split the data into training and test data, and use the training data to assign mass functions.
- Assign mass values for each attribute based on the built mass functions. This is done three times, first for the tree model, then for the animal model, then for the lightning model.
- For each record, combine the mass values of all the attributes using the tree model.
- For each record, combine the mass values of all the attributes using the animal model.
- For each record, combine the mass values of all the attributes using the lightning model.
- For each record, combine the mass values for the tree, animal and lightning models.
- Use a decision rule to identify the cause of the power outage.

The likelihood measure is used when assigning mass values as [20] and [21] have shown that this provides useful information for outage fault identification when used with other methods. The likelihood is a value between zero and one that gives the probability of a fault happening under a specific event, and is defined as:

$$L_{ij} = \frac{N_{ij}}{N_j}, \qquad (23)$$

where $L_{ij}$ is the likelihood measure of fault $i$ given event $j$, $N_{ij}$ is the number of outages caused by fault $i$ under event $j$, and $N_j$ is the total number of outages under event $j$. Here, $i \in \Theta$ = {tree, animal, lightning, others} and $j$ is the set of values that each attribute can take. Both $N_{ij}$ and $N_j$ are calculated from the training data, so that the general likelihoods of tree-fault, lightning-fault and animal-fault outages can be deduced for each attribute, and used to form the specific mass functions. These can be represented mathematically as:

$$\begin{aligned}
&m(\Theta) = 1, \ L_{ij} = L_i, \\
&m(i) = \frac{(L_{ij} - L_i)}{1 - L_i}, m(\Theta) = 1 - m(i), \ L_{ij} > L_i, \\
&m(\neg i) = \frac{(L_i - L_{ij})}{L_i}, m(\Theta) = 1 - m(\neg i), \ L_{ij} < L_i,
\end{aligned} \qquad (24)$$

and can be explained in the following way. For each attribute of each outage, if the likelihood under event $j$ for fault $i$ is larger than the general likelihood for fault $i$, then the outage is more likely caused by fault $i$ under event $j$ for that attribute. Here, $j$ is a specific value of that attribute, for example 'rain' for the attribute WE. However, if the likelihood is smaller than the general likelihood, then the outage is not likely caused by fault $i$ under event $j$ for that attribute, it is more likely cased by fault 'not $i$'. If the likelihood is equal to the general likelihood, then no judgement is made based on that information, i.e., the uncertainty is equal to 1. The chosen mass function structure ensures that conditions (3) – (5) are met, see Section 2.

Since the DOD dataset is imbalanced, classification accuracy is not a suitable performance measure in this case. For example, if 97% of data items are actually due to other causes and only 3% are due to animals, and all faults are classed as 'other', then this gives a classification accuracy of 97%, which is misleading. For this reason, the geometric mean of accuracies (*g*-mean), introduced by Kubat *et al.* for imbalanced datasets [27] is used instead. The *g*-mean is given by:

$$g - mean = \sqrt{Acc^+ \times Acc^-}, \tag{25}$$

$$Acc^+ = \frac{TP}{TP + FN}, \tag{26}$$

$$Acc^- = \frac{TN}{TN + FP}, \tag{27}$$

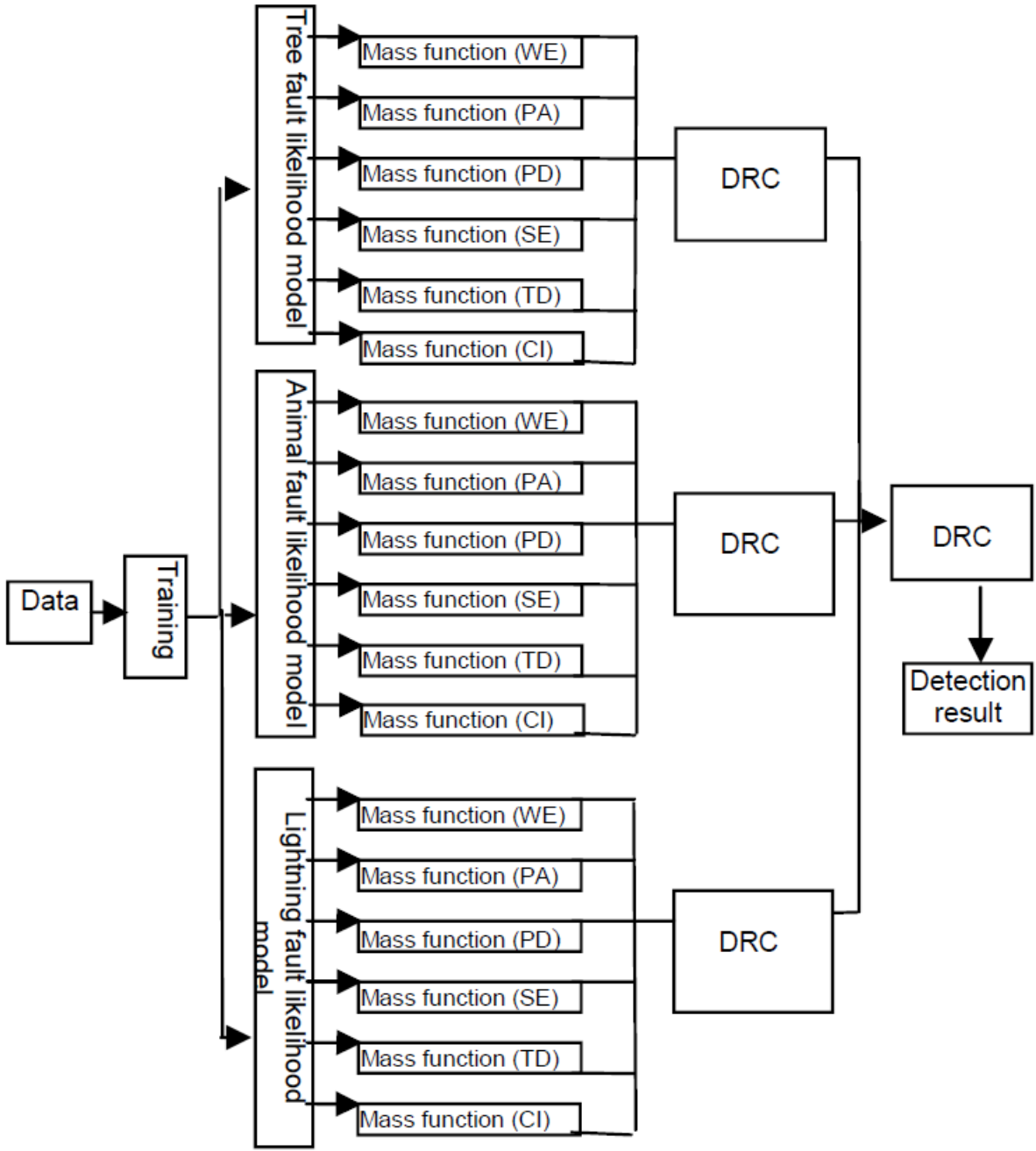

Fig. 8. System flowchart for classification of the Duke Outage dataset.

where $Acc^+$ and $Acc^-$ are the true positive and true negative rates respectively, and *TP*, *TN*, *FP*, and *FN* are the numbers of true positives, true negatives, false positives and false negatives respectively. Since all these metrics deal with binary classification (either positive or negative), the *g*-mean, $Acc^+$ and $Acc^-$ must be measured separately for each class; tree, animal and lightning. So, for example, $Acc^+_{tree}$ is the true positive rate for the class tree, and is derived from $TP_{tree}$, the number of correctly classified tree-caused faults and $FN_{tree}$, the number of falsely classified tree-caused faults. Similarly, $Acc^-_{lightning}$ is the true negative rate for the class lightning, and is derived from $TN_{lightning}$, the number of correctly classified non-lightning-caused faults and $FP_{lightning}$, the number of falsely classified non-lightning-caused faults. The $g$-mean gives an overall evaluation of $Acc^+$ and $Acc^-$, exhibiting a high value if both are large and the difference between them is small.

The dataset is divided into two subsets, 1994 - 1999 data and 2000 - 2002 data, with the former serving as training data and the latter as testing data. Initially, three different types of experiment are carried out; first all the attributes are used, then each of the single attributes, and finally, all the attributes except one are combined. In these experiments the same attribute combination is used for each model. In a fourth set of trials different combinations are tried with each model (i.e. those that proved best in the initial three sets of trials). The system is deterministic once the attribute combinations are chosen (since the training set is fixed), so each experimental scenario is run only once.

### 6.3. Experimental results using the Duke Outage dataset

Table 7 shows the $g$-mean, $Acc^+$ and $Acc^-$ values achieved for the tree class after the final DRC step, using the various attribute combinations trialled. The final six rows present the data when different attribute combinations are used for each individual model. Tables 8 and 9 reveal the same statistics for the animal and lightning classes respectively. The $g$-mean, $Acc^+$, and $Acc^-$ values are also shown graphically in Figures 9 - 11.

Figures 9 - 11 and Tables 7 – 9 show that the results using all six attributes are quite poor, with few values exceeding 0.50. This is probably because some of the attributes are very weak classifiers and their inclusion in the combined system seems to have a marked effect on overall performance. Indeed, the single-attribute results demonstrate this. In the case of the animal class for example, attribute 2 has an extremely low true positive rate (0.045), which is reflected in the true positive rate (0.112) when all the attributes are used. However, when attribute 2 is excluded the true positive rate rises to 0.731.

The performance of the single attributes is not consistent throughout the classes, for example, attribute 2 is very poor in the animal class ($g$-mean 0.164), but performs much better for the lightning class with a $g$-mean of 0.500. Results using single attributes are also generally quite poor with none having a $g$-mean above 0.50, and some in the lightning class having a zero $g$-mean. The $g$-means generally improve when five attributes are used, especially for the lightning and tree classes, but are still quite low.

The best results are obtained when different attribute combinations are used for the different classes, for example using attribute triplets of 2, 4 and 6 for the tree class, 1, 3 and 4 for the animal class and 1, 2 and 5 for the lightning class yields $g$-means of 0.530, 0.603 and 0.504 respectively. When the tree class excludes attribute 5, the animal class excludes attribute 2 and the lightning class uses attribute 2 only the $g$-means are 0.544, 0.614, and 0.539 respectively. This represents the best performance overall for the D-S classifier.

**Table 7 Results for the tree class.**

| Attributes | g-mean | $Acc^+$ | $Acc^-$ |
|---|---|---|---|
| All | 0.494 | 0.468 | 0.522 |
| 1 only | 0.280 | 0.384 | 0.204 |
| 2 only | 0.439 | 0.365 | 0.529 |
| 3 only | 0.306 | 0.202 | 0.463 |
| 4 only | 0.380 | 0.328 | 0.440 |
| 5 only | 0.165 | 0.160 | 0.170 |
| 6 only | 0.383 | 0.683 | 0.215 |
| ¬1 | 0.498 | 0.481 | 0.514 |
| ¬2 | 0.487 | 0.445 | 0.533 |
| ¬3 | 0.492 | 0.466 | 0.519 |
| ¬4 | 0.462 | 0.421 | 0.508 |
| ¬5 | 0.505 | 0.488 | 0.523 |
| ¬6 | 0.498 | 0.482 | 0.514 |
| 234 & 134 & 125 | 0.530 | 0.456 | 0.617 |
| 234 & 346 & 125 | 0.525 | 0.442 | 0.623 |
| 246 & 134 & 125 | 0.530 | 0.463 | 0.608 |
| 246 & 346 & 125 | 0.532 | 0.465 | 0.609 |
| ¬5 & ¬2 & ¬3 | 0.549 | 0.483 | 0.625 |
| ¬5 & ¬2 & 2 | 0.544 | 0.505 | 0.608 |

**Table 8 Results for the animal class**

| Attributes | g-mean | $Acc^+$ | $Acc^-$ |
|---|---|---|---|
| All | 0.262 | 0.112 | 0.615 |
| 1 only | 0.312 | 0.472 | 0.206 |
| 2 only | 0.164 | 0.045 | 0.595 |
| 3 only | 0.450 | 0.841 | 0.241 |
| 4 only | 0.483 | 0.853 | 0.274 |
| 5 only | 0.200 | 0.343 | 0.116 |
| 6 only | 0.437 | 0.640 | 0.298 |
| ¬1 | 0.270 | 0.119 | 0.612 |
| ¬2 | 0.566 | 0.731 | 0.438 |
| ¬3 | 0.221 | 0.079 | 0.621 |
| ¬4 | 0.174 | 0.051 | 0.600 |
| ¬5 | 0.287 | 0.133 | 0.619 |
| ¬6 | 0.248 | 0.100 | 0.618 |
| 234 & 134 & 125 | 0.620 | 0.759 | 0.506 |
| 234 & 346 & 125 | 0.617 | 0.748 | 0.508 |
| 246 & 134 & 125 | 0.603 | 0.701 | 0.518 |
| 246 & 346 & 125 | 0.611 | 0.731 | 0.511 |
| ¬5 & ¬2 & ¬3 | 0.609 | 0.675 | 0.549 |
| ¬5 & ¬2 & 2 | 0.614 | 0.703 | 0.536 |

**Table 9 Results for the lightning class**

| Attributes | g-mean | $Acc^+$ | $Acc^-$ |
|---|---|---|---|
| All | 0.455 | 0.407 | 0.508 |
| 1 only | 0.308 | 0.363 | 0.261 |
| 2 only | 0.500 | 0.531 | 0.471 |
| 3 only | 0.000 | 0.000 | 0.391 |
| 4 only | 0.219 | 0.115 | 0.415 |
| 5 only | 0.336 | 0.832 | 0.136 |
| 6 only | 0.000 | 0.000 | 0.391 |
| ¬1 | 0.473 | 0.442 | 0.506 |
| ¬2 | 0.344 | 0.230 | 0.515 |
| ¬3 | 0.477 | 0.451 | 0.503 |
| ¬4 | 0.452 | 0.425 | 0.481 |
| ¬5 | 0.472 | 0.434 | 0.515 |
| ¬6 | 0.455 | 0.407 | 0.508 |
| 234 & 134 & 125 | 0.501 | 0.442 | 0.568 |
| 234 & 346 & 125 | 0.501 | 0.442 | 0.567 |
| 246 & 134 & 125 | 0.504 | 0.451 | 0.563 |
| 246 & 346 & 125 | 0.500 | 0.442 | 0.565 |
| ¬5 & ¬2 & ¬3 | 0.503 | 0.434 | 0.584 |
| ¬5 & ¬2 & 2 | 0.539 | 0.504 | 0.576 |

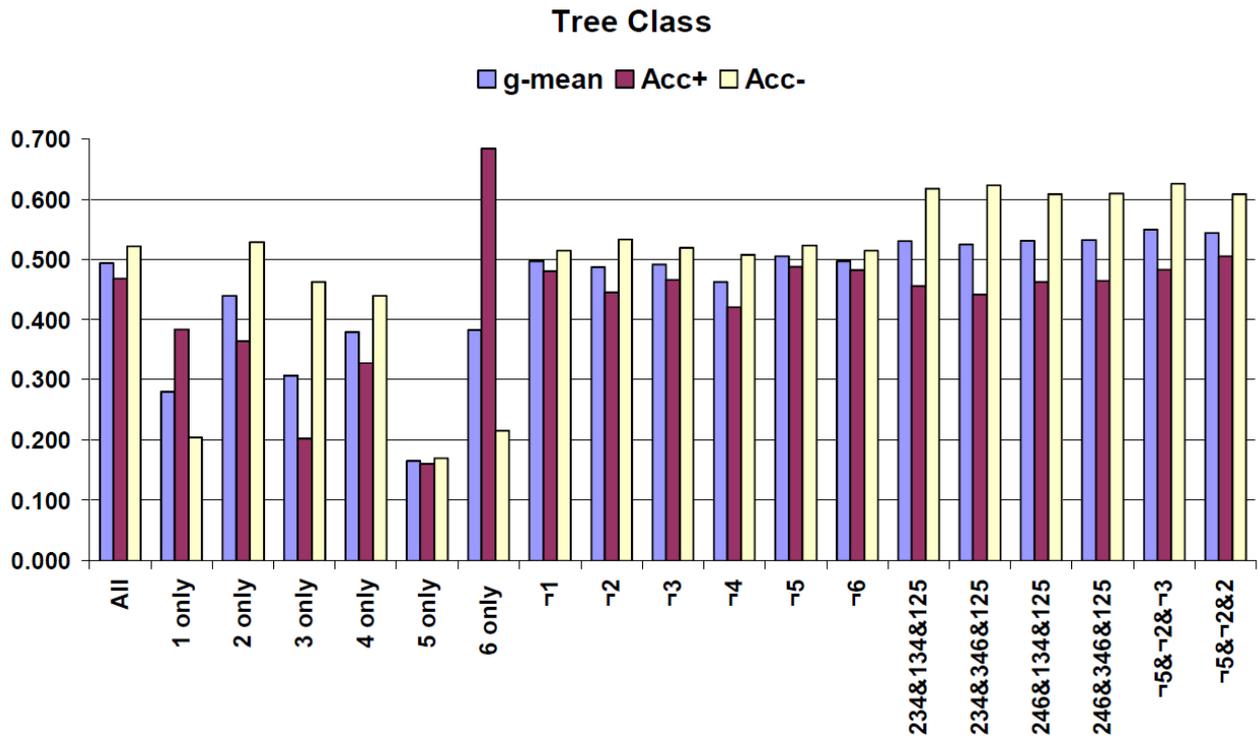

Fig. 9. Histogram of performance statistics for the tree class, showing use of all attributes, single attributes, five attributes, and different combinations of attributes for the different models.

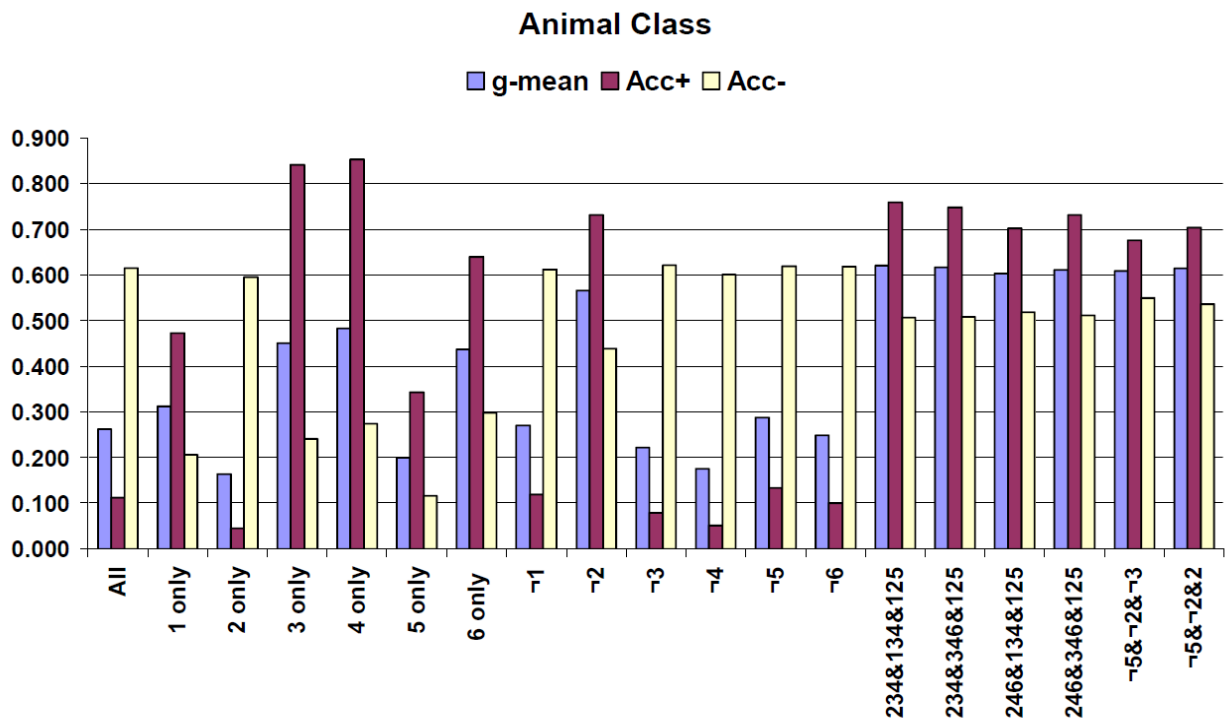

Fig. 10. Histogram of performance statistics for the animal class, showing use of all attributes, single attributes, five attributes, and different combinations of attributes for the different models.

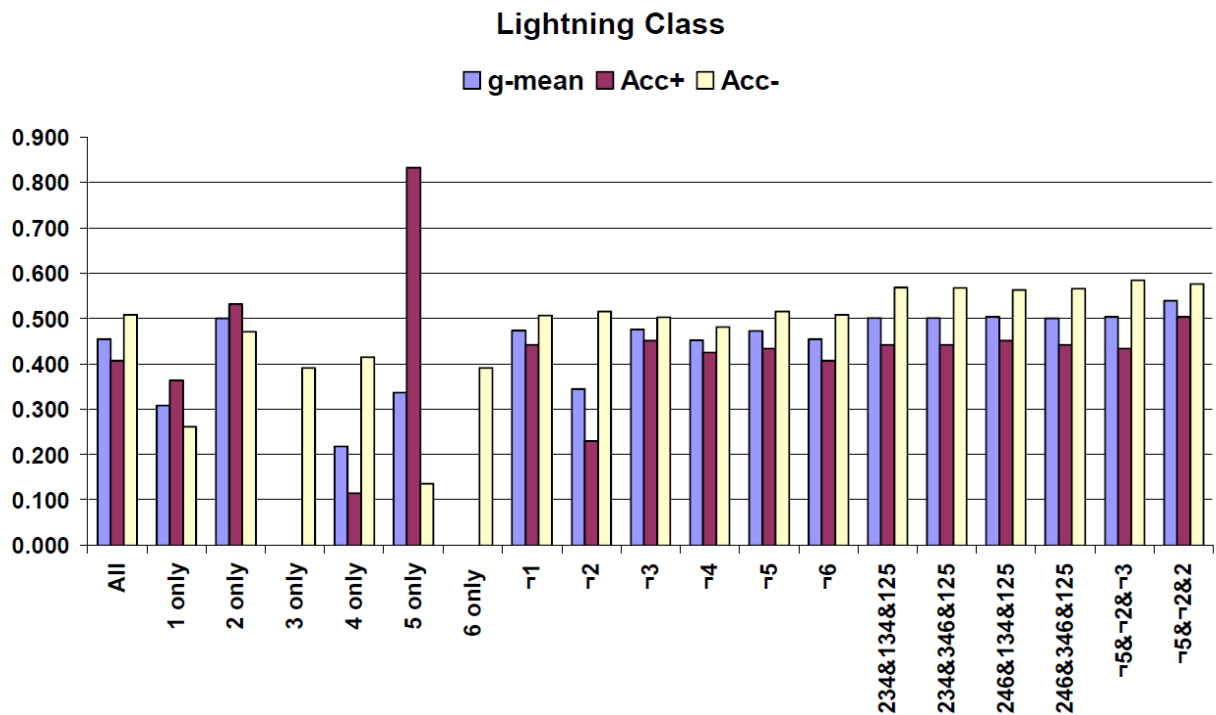

**Fig. 11.** Histogram of performance statistics for the lightning class, showing use of all attributes, single attributes, five attributes, and different combinations of attributes for the different models.

These results collectively show that the system needs to be run using different combinations of attributes for each class in order to perform well. They also provide a stark contrast to the WBCD results, which tend to increase in accuracy when more attributes are used, despite some attributes being poorer.

Table 10 shows the calculated $g$-mean values for animal, tree and lightning-caused faults when the same data is classified using the Naive Bayes approach and the algorithms created by Xu *et al*. The latter consist of an Artificial Neural Network (ANN) [20], an Artificial Immune Recognition System (AIRS) [18], a fuzzy algorithm called the E-Algorithm [19] and a Fuzzy Artificial Immune Recognition System (FAIRS) [25]). FAIRS is a hybrid system that incorporates the quick searching capability and memory mechanism of the AIRS algorithm and the inference rule extracting capability of the E-algorithm. These algorithms also use the 1994 - 1999 data as training data and the 2000 - 2002 data as test data. The $g$-means for the best D-S system (i.e. excluding attributes 5 and 2 for the tree and animal classes respectively, and using attribute 2 only for the lightning class) are shown in the table for comparison.

**Table 10 A comparison with other systems**

| SYSTEM | $g$-mean Tree | Animal | Lightning |
|---|---|---|---|
| ANN | 0.56 | 0.58 | 0.23 |
| E-Algorithm | 0.32 | 0.52 | **0.70** |
| AIRS | 0.57 | **0.72** | 0.57 |
| FAIRS | 0.59 | 0.65 | **0.70** |
| D-S (Best) | 0.54 | 0.61 | 0.54 |
| Naive Bayes | **0.62** | 0.71 | 0.64 |

Table 10 shows that the D-S system is competitive with the other algorithms; in particular it outperforms two of them, ANN and the E-algorithm since these are both unable to achieve $g$-means above 0.5 for all classes. The system is bettered by the AIRS and FAIRS algorithms, which both have a higher $g$-mean for each class. The D-S system is rather let down by its performance for the lightning class, which is probably due to the small percentage of faults that can be attributed to this class.

The D-S results are adequate given that this is a difficult dataset to classify; the system's achievement of a $g$-mean above 0.50 for each class is important and is indicative of the suitability of the D-S framework for creating classifiers to deal with inconsistent, non-numerical data. However, the experiments have revealed that careful selection of the attributes for each individual class is necessary in order to achieve good results.

Since the results are promising and reasonably close to the FAIRS algorithm it is worth attempting to optimize the mass functions used in order to investigate whether any improvement can be achieved. This is the subject of the next section.

### 6.4. Optimization of the mass functions and results

It is possible to optimize the mass functions (24) using simulated annealing (SA), see [29]. This process is concerned with finding parameter values $\alpha$ and $\beta$ between 1 and 10 in (28) that give the highest $g$-mean product for the three classes; tree, animal and lightning. The updated mass functions are hence given by:

$$
\begin{aligned}
&m(\Theta) = 1, \; L_{ij} = L_i, \\
&m(i) = \frac{\alpha(L_{ij} - L_i)}{1 - L_i}, \; m(\Theta) = 1 - m(i), \; L_{ij} > L_i, \; \frac{\alpha(L_{ij} - L_i)}{1 - L_i} \leq 1, \\
&m(i) = 1, \; m(\Theta) = 1 - m(i), \; L_{ij} > L_i, \; \frac{\alpha(L_{ij} - L_i)}{1 - L_i} > 1, \\
&m(\neg i) = \frac{\beta(L_i - L_{ij})}{L_i}, \; m(\Theta) = 1 - m(\neg i), \; L_{ij} < L_i, \; \frac{\beta(L_i - L_{ij})}{L_i} \leq 1, \\
&m(\neg i) = 1, \; m(\Theta) = 1 - m(\neg i), \; L_{ij} < L_i, \; \frac{\beta(L_i - L_{ij})}{L_i} > 1.
\end{aligned}
\qquad (28)
$$

The initial value is set at 10 for both $\alpha$ and $\beta$ and this is decremented upon every SA iteration by multiplying by a factor of 0.99. Full details of the SA process and pseudo code for generating new $\alpha$ and $\beta$ values are provided in [28]. Only the best D-S system from the non-optimized results (i.e. ¬5 ¬2 and 2) is used in the optimization process.

The results show an optimum $\alpha$ value of 2.6 and an optimum $\beta$ value of 1.3. These produce $g$-mean values of 0.57, 0.63, and 0.54 for the tree, animal and lightning classes respectively. This represents a slight improvement for the tree and animal classes, but no improvement for the lightning class.

### 7. Conclusions

This work has utilized the D-S theory (in particular mass functions and DRC) as a framework for creating classification algorithms, and has applied them to three standard benchmark datasets, the WBCD dataset, the Iris dataset, and part of the Duke Outage dataset. For the WBCD, the mass functions were created by considering threshold values in the training data and using a sigmoid model. In this case, classification was a simple one-step process. The accuracy proved to be much higher when all the data attributes were considered (97.6%), and this result was superior to other published results for other popular methods. Furthermore, the D-S method permitted the inclusion of data items that contained missing values in the dataset. Some of the other methods were unable to do this.

The Iris dataset was grouped into its three classes using a two-step process that made use of the class boundary values in the training data for each attribute. Data items that were not recognized as belonging to a single class were subject to a further classification step that utilized standard deviation measures to select an attribute on which final classification was based. A mean classification accuracy of 95.47% was achieved, which was in the same range as existing results using other methods.

A system was also built to tackle the Duke Outage dataset, which has proved difficult to classify accurately in the past. The system initially performed binary classification on each class separately and then combined the mass values for each class using DRC, so that the data was grouped into its four classes. In this case, mass functions were based on likelihood measurements from the training data. Results showed that the system performed poorly when the same attribute combinations were used with each class, but classified reasonably well compared with other published state-of-the-art methods (in fact better than two of them), when different attribute combinations were used for each class. A simulated annealing method of optimization was trialled on the mass functions used for this dataset, but produced only a slight improvement in $g$-mean scores for the tree and animal classes. No improvement was made to the lightning class $g$-mean.

This paper has hence demonstrated that the D-S approach works well with all three datasets, provided the system is designed in the right way and the attributes are carefully selected. Attribute selection appears to influence overall performance considerably, for example, use of all the attributes worked well for the WBCD but not for the Duke Outage data.

However, since there is much scope for system and mass function design, and the designs themselves are very data-dependent it is not possible to make generalized statements about D-S as a single, rigid, classification system. The D-S theory provides the *framework* for system design only, and in this sense allows the creation of systems that can be essentially tailored towards the specific problem domain of interest. This may be considered a disadvantage in that there are no strict guidelines for the detailed design of such systems, but it may also be thought of as an advantage, since the flexibility allows for the tweaking and refinement of the system until the desired output levels are reached, especially if this refinement process can be automated in some way. In particular, automating the attribute selection and mass function calculation processes may make the Dempster-Shafer approach an objective and accurate replacement for current state of the art classification systems.